\documentclass[]{ceurart}

\usepackage{listings}
\usepackage{xcolor}
\begin{document}

\copyrightyear{2024}
\copyrightclause{Copyright for this paper by its authors.
  Use permitted under Creative Commons License Attribution 4.0
  International (CC BY 4.0).}

\conference{CHR 2024: Computational Humanities Research Conference, December 4–6, 2024, Aarhus, Denmark}

\title{SCIENCE IS EXPLORATION: Computational Frontiers for Conceptual Metaphor Theory}

\author[1,2]{Rebecca M. M. Hicke}[%
orcid=0009-0006-2074-8376,
email=rmh327@cornell.edu,
url=https://rmatouschekh.github.io,
]
\cormark[1]
\address[1]{Department of Computer Science, Cornell University, USA}
\address[2]{Center for Humanities Computing, Aarhus University, Denmark}

\author[2,3]{Ross Deans Kristensen-McLachlan}[%
orcid=0000-0001-8714-1911,
email=rdkm@cc.au.dk,
]
\address[3]{Department of Linguistics, Cognitive Science, and Semiotics, Aarhus University, Denmark}

\cortext[1]{Corresponding author.}

\begin{abstract}
Metaphors are everywhere. They appear extensively across all domains of natural language, from the most sophisticated poetry to seemingly dry academic prose. A significant body of research in the cognitive science of language argues for the existence of \textit{conceptual metaphors}, the systematic structuring of one domain of experience in the language of another. Conceptual metaphors are not simply rhetorical flourishes but are crucial evidence of the role of analogical reasoning in human cognition. In this paper, we ask whether Large Language Models (LLMs) can accurately identify and explain the presence of such conceptual metaphors in natural language data. Using a novel prompting technique based on metaphor annotation guidelines, we demonstrate that LLMs are a promising tool for large-scale computational research on conceptual metaphors. Further, we show that LLMs are able to apply procedural guidelines designed for human annotators, displaying a surprising depth of linguistic knowledge.
\end{abstract}

\begin{keywords}
  conceptual metaphor theory \sep
  large language models \sep
  pragglejaz
\end{keywords}

\maketitle
\newcommand{\epigraph}[3]{\begin{quotation} \textit{#1} \end{quotation} \begin{flushright} - #2 \textit{#3}\end{flushright} }

\epigraph{"Metaphors are much more tenacious than facts."}{Paul de Man \cite{deman_1973}}{}

\section{Introduction}

Metaphor is commonly understood as the description of one concept in the vocabulary of another, typically for poetic, rhetorical, or otherwise literary effect. However, some metaphors are so systematic and conventionalized that we barely recognize them as such. Consider, for example, the commonplace metaphor \textsc{life is a journey}. While this is certainly a metaphor in its own right, it can also be instantiated in ways which do not directly refer to either life or journeys: "We have \textit{come a long way}"; "I'm \textit{going through a rough patch}"; "She's at a \textit{crossroads}." A metaphor like \textsc{life is a journey} is more than simple figurative language. Instead, it allows us to make sense of the process of a normal human life through the rich language of journeys and voyages. These kinds of systematic mappings are known as \textit{conceptual metaphors} \cite{lakoff-johnson-1980}.

Conceptual metaphors are so deeply ingrained in our everyday language that the most frequent usage of a word or phrase may be metaphorical. This can make it difficult to recognize the presence of metaphorical language, even for linguists and other specialists \cite{group2007mip}. Moreover, processing and understanding conceptual metaphors necessarily involves encyclopedic general knowledge of the world, as well as shared cultural norms and stereotypes between speaker/hearer or writer/reader. Identifying conceptual metaphors has therefore consistently proven to be a challenge for both natural language processing (NLP) and artificial intelligence (AI). Given that contemporary transformer-based models `learn' by consuming huge amounts of natural language and build knowledge by looking at how words are commonly used, it seems that they should similarly struggle to recognize these metaphors.

Nevertheless, being able to consistently identify the presence of conceptual metaphors in discourse would be greatly beneficial for fields such as computational literary studies and cultural analytics more broadly. In this paper, we therefore set out to test empirically whether LLMs can be used productively in the context of conceptual metaphor theory (CMT). If the answer is no, it may be evidence that developing computational approaches to conceptual metaphor remains an intractable problem. If, on the other hand, LLMs \textit{can} consistently identify the presence of conceptual metaphors,  the question then arises of exactly how and where LLMs encode metaphoricity.

\section{Related Work}

\subsection*{Cognitive Linguistic Foundations}
The study of CMT was inaugurated in the 1980s in \citet{lakoff-johnson-1980}. Central to the argument in \cite{lakoff-johnson-1980} is the fundamental role that metaphor plays in structuring everyday discourse. Rather than simply being a literary or rhetorical flourish, Lakoff and Johnson argue that conceptual metaphors structure our common understanding of entities, processes, or entire semantic domains. CMT played a crucial role in the development of non-generativist cognitive linguistics \cite{croftcruse2004, langackervol1_1987,langackervol2_1991, talmy_vol1_2000, talmy_vol2_2000}. An important development came with the introduction of \textit{conceptual blending} (otherwise known as \textit{conceptual integration}), in which the kinds of analogical reasoning underlying conceptual metaphors are argued to be more fundamental to human language and cognition \cite{FauconnierGilles1994MSAo, blending_2002, DancygierBarbara2005MSiG}. While there are dissenting voices \cite{GibbsRaymondW.2009WDSP}, recent decades have seen a growing body of empirical research which claims to provide experimental evidence in favour of CMT \cite{BoroditskyLera2000Msut, CasasantoDaniel2008Titm, ThibodeauPaulH.2017HLMS}.

\subsection*{Metaphor and Artificial Intelligence}
The extensive role and function of metaphor in natural language has not escaped researchers in NLP. Indeed, a great deal of early work in AI concerned exactly how a computer could make sense of metaphorical language \cite{CarbonellJaimeG1981MAIP, russell-1976-computer, RussellSylviaWeber1986IaEi} (see \cite{BarndenJohnA.2008MaAI} for a comprehensive overview). Much of this work relied on so-called symbolic AI, but struggled because GOFAI (good, old-fashioned AI) could not provide the analogous reasoning skills and extensive general knowledge required to process metaphorical language. In the context of conceptual blending specifically, there have been tentative attempts to model the underlying processes involved computationally \cite{EPPE2018105}. However, the encyclopedic world knowledge required in CMT is challenging to encode since the domain is essentially unbounded. Moreover, a core issue at the heart of computational metaphor analysis has historically been the lack of a shared definition of what is being studied, along with the scarcity of robust data and evaluation strategies \cite{shutova2015design}.

\subsection*{Recent Developments}
However, recent years have again seen a growing interest in quantitative and computational approaches to metaphor, primarily in more applied fields such as corpus linguistics and stylistics \cite{deignansemino2010, krennmayr-2015-what, SeminoElena2018MCat}. With the advent of neural language models, research in NLP has increasingly attempted to discover if (and how) word embeddings capture aspects of metaphoricity \cite{mao-etal-2018-word, tong-etal-2021-recent, panicheva-2023-towards, ptivcek2023methods}. More recently, work has been done on explicitly probing pre-trained language models to inspect how metaphorical knowledge is encoded \cite{aghazadeh-etal-2022-metaphors} and creating shared task datasets \cite{tong-etal-2021-recent}. In the context of LLMs specifically, GPT-3 has been shown to perform reasonably well at predicting source domains for English metaphors, although it performs significantly worse for Spanish \cite{wachowiak-gromann-2023-gpt}. These are promising developments and point to the increasing awareness of the practical and theoretical challenges underlying the nature of figurative language in NLP. To date, though, we are unaware of existing research which seeks to operationalize existing metaphor identification procedures for LLMs.

\section{Methods}

\begin{figure}
    \centering
    \includegraphics[scale=0.6]{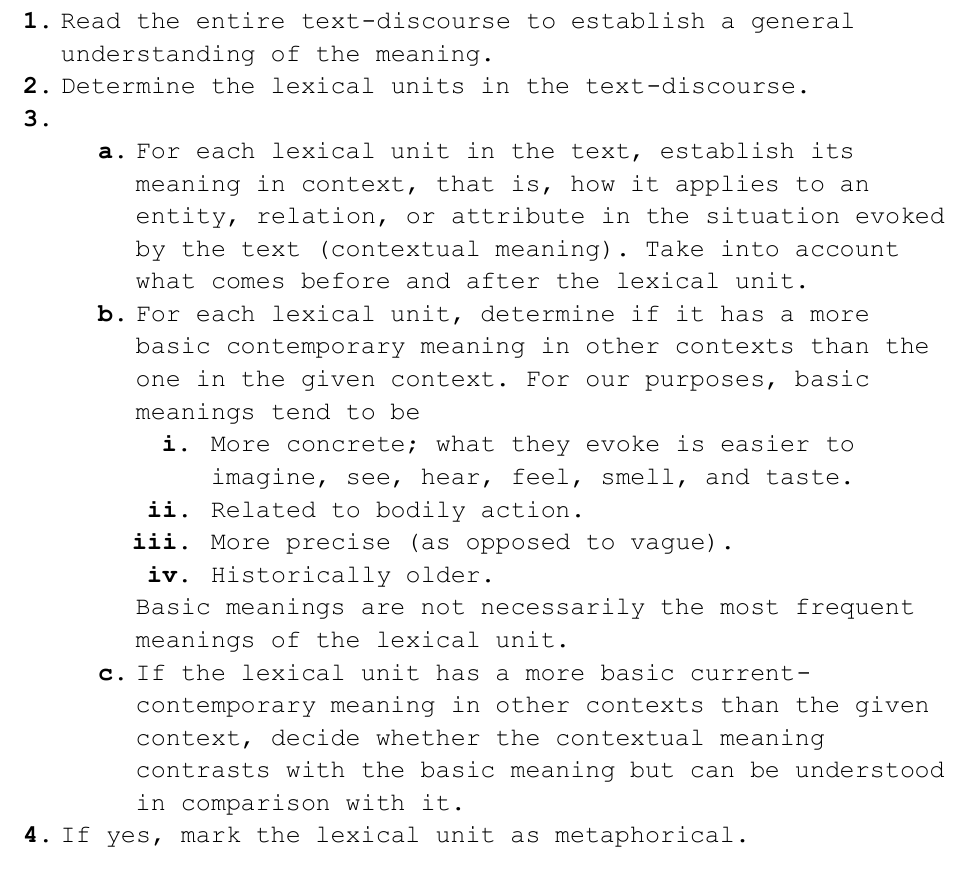}
    \caption{The metaphor identification procedure (MIP) introduced by the Pragglejaz Group in \cite{group2007mip}.}
    \label{fig:mip-procedure}
\end{figure}

\begin{figure}
    \centering
    \includegraphics[scale=0.6]{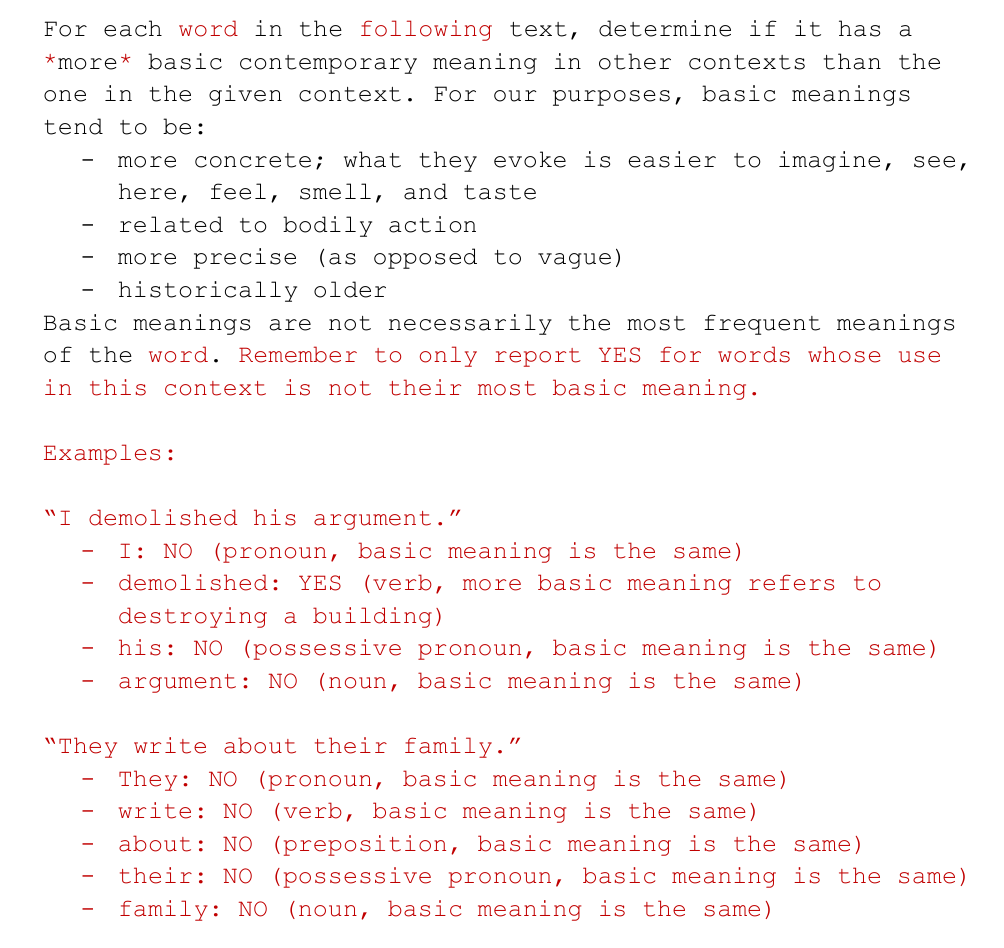}
    \caption{The modified prompt used in all experiments. All significant (non-formatting) changes from Step 3b of the MIP procedure are highlighted in red. The sentence to be analyzed is appended in quotations to the bottom of the prompt when querying models.}
    \label{fig:model-prompt}
\end{figure}

We are interested in studying whether LLMs are capable of leveraging the contextual and cultural knowledge necessary to perform difficult metaphor identification tasks, such as recognizing conceptual metaphors. To do this, we operationalize the Metaphor Identification Procedure (MIP), a set of annotation guidelines introduced by the Pragglejaz Group in 2007 \cite{group2007mip}. The steps of MIP are detailed in Figure \ref{fig:mip-procedure}. MIP and its variants are among the most commonly used procedures for metaphor annotation in work in corpus linguistics and stylistics; we therefore take it as a reasonable proxy for the metaphor identification process in humans. 

Examining MIP, the first few steps have clear parallels in transformer-based LLMs. Step 1 is performed by the attention mechanism. Creating the representation of each word, in traditional attention mechanisms, requires `looking' at every other word in the text, thus allowing the model to ``establish a general understanding of [its] meaning.'' Step 2, determining the lexical units in a text, is directly paralleled by tokenization. Finally, Step 3a is achieved via contextual embeddings, whose explicit purpose is to represent the specific meaning of a word in context. However, Step 3b of MIP has no clear counterpart in transformer-based models. This is because the `basic meaning' of a lexical unit, as defined by MIP, is not obviously stored or represented within these models.

We want to know if state-of-the-art LLMs are capable of replicating Step 3b of the MIP procedure despite these barriers, so we transform it into a prompt for instruction-tuned models (Figure \ref{fig:model-prompt}). In creating the prompt, we leave the text of Step 3b almost entirely intact. ``Lexical units'' are replaced by ``words'' in order to encourage consistent outputs. Emphasis is added in several locations to only annotating words which have a \textit{more} basic meaning than their usage in the input text. Finally, two examples are appended to the prompt. These examples are meant to facilitate in-context learning and provide a structure for the models to mimic in their outputs. They instruct the model to provide a parenthetical expression next to each word consisting of the word's part of speech and its more basic meaning, if one exists. These expressions were introduced by the \texttt{gpt-4o} model itself during initial exploration and were then included in the prompt because they force the model to provide explanations for its annotations. This allows for further evaluation of the outputs and ultimately means they play a similar role to Chain of Thought prompts, which have been shown to improve model performance across a variety of tasks \cite{wei2024chainofthought}.

We use this prompt to query three models from OpenAI's GPT family, chosen because of their size, popularity, and high performance on a wide range of tasks. Specifically, we look at \texttt{gpt-3.5-turbo},\footnote{\texttt{gpt-3.5-turbo-0125}} \texttt{gpt-4-turbo},\footnote{\texttt{gpt-4-turbo-2024-04-09}} and \texttt{gpt-4o}.\footnote{\texttt{gpt-4o-2024-05-13}} We access each model using OpenAI's Chat Completion API. A system prompt (\textit{``You are a helpful assistant. You have extensive linguistic knowledge.''}) is provided each time the model is queried. Then, the main prompt (Figure \ref{fig:model-prompt}) is passed as a user message, with the current text of interest appended in quotations at the end. All parameters except nuclear sampling (\texttt{top\_p}) are left at their default values during prompting for all models. 
We set \texttt{top\_p} to 0.1, which means that the models only select from the tokens which make up the top 10\% of the probability mass when choosing each next token. We adjust the \texttt{top\_p} value because we are not prompting for diverse or creative text, but instead for reliable and accurate outputs, which we hypothesize will be more likely produced from high-probability tokens. The cost of all experiments using the OpenAI API was $\sim$\$100.

\section{Data}

We evaluate each model on two datasets. First, we wish to study the models' basic ability to use MIP to correctly identify metaphorical words. To this end, we use the Trope Finder (TroFi) dataset \cite{birke-sarkar-2006-clustering}, which consists of sentences drawn from Wall Street Journal issues published between 1987 and 1989. Each sentence contains one of a list of fifty words whose usage is annotated as literal or non-literal. We only consider sentences from the dataset that have been labeled by humans and take their annotations as ground truth, excluding all sentences labeled only by the TroFi clustering system. This leaves us with 3,736 sentences. Each model is then prompted as described above for each sentence and evaluated only on the words annotated in TroFi. A response is marked as correct if the model says ``YES'' there is a more basic meaning for the word of interest and the label from TroFi is ``literal'' and vice versa. Occasionally, the models provide labels for a phrase in the input sentence instead of individual words; the label provided for the entire phrase is then applied to the word of interest if necessary. If a model does not provide a label for a word, we act as if the incorrect label was given.

However, the TroFi dataset does not permit for a very fine-grained analysis of the models' capabilities. In order to facilitate such an analysis, we create a new dataset consisting of example sentences pulled from Lakoff and Johnson's original work on CMT \cite{lakoff-johnson-1980}. 
We collect all full sentence examples from the text that we believe can be understood as metaphorical without context, leaving us with 544 sentences, which we refer to as the \textit{MWLB dataset}. The dataset is made publicly available, along with all model outputs and human annotations.\footnote{\url{https://github.com/rmatouschekh/science-is-exploration}}

Each model is again prompted to provide word-level annotations for every sentence. Then, we perform detailed qualitative analysis on each model's output for a subset of 100 sentences. Specifically, we evaluate each output on five binary categories:

\begin{itemize}
    \item \textbf{L\&J Metaphor(s) -- Identified}: Whether the model has correctly identified \textit{all} metaphors highlighted by Lakoff and Johnson. For metaphorical phrases, the model's response is considered correct if it has identified at at least one key word as having a more basic meaning.
    \item \textbf{L\&J Metaphor(s) -- Correct Basic Meanings:} Whether the model has provided a correct, or plausible, more basic meaning for \textit{all} metaphorical words identified in the category above. A label of 1 is applied only if all metaphors from Lakoff and Johnson have been correctly identified.
    \item \textbf{Additional Annotations:} Whether the model has labeled any words not highlighted by Lakoff and Johnson as having a more basic meaning.
    \item \textbf{Additional Annotations -- Metaphorical:} Whether \textit{all} the additional words labeled by the model are plausibly metaphorical or may have a more basic meaning.
    \item \textbf{Additional Annotations -- Correct Basic Meanings:} Whether the model has provided a correct, or plausible, basic meaning for \textit{all} additional labeled words. It is possible for models to give the correct basic meaning for a word even if it is not plausibly metaphorical. 
\end{itemize}

\begin{table*}[t]
    \caption{The precision and recall values for the identification of literal and metaphorical word usage in the TroFi dataset by all models.}
    \label{tab:trofi-values}
    \begin{tabular}{lcccc}
    \toprule
         & \multicolumn{2}{c}{\textbf{Metaphorical}} & \multicolumn{2}{c}{\textbf{Literal}} \\
        \cmidrule(l){2-3} \cmidrule(l){4-5}
        & Precision & Recall & Precision & Recall \\
        \midrule
        \texttt{3.5-turbo} & 58.30 & 97.90 & 66.42 & 5.59 \\
        \texttt{4-turbo} & 74.80 & 86.90 & 77.41 & 60.53 \\
        \texttt{4o} & 73.40 & 93.66 & 83.69 & 54.24 \\
        \bottomrule
    \end{tabular}
\end{table*}

All annotations are performed by one of the authors, using MIP as a guideline. Confusing or difficult annotations are then discussed between both authors till a consensus is reached. These annotations are necessarily subjective, and we do not claim that they represent a ground truth. Instead, since these evaluations are informed by the same procedure used to create the outputs, we claim that they allow us to study the \textit{plausibility} of the models' performance, if not strictly their \textit{accuracy}. In future work, further evaluation of model outputs with multiple trained annotators may allow for a more concrete analysis of model performance.

\section{Results and Discussion}

On the TroFi dataset, all models demonstrate an ability to distinguish between literal and non-literal word usage with generally better than chance performance (Table \ref{tab:trofi-values}). Thus, they seem to be able to apply the MIP procedure for identifying metaphors and determine computationally whether a word has a more basic meaning. This is surprising since, as discussed above, basic word meanings are not necessarily the most frequent. Nothing about the language modeling approach to pre-training models guarantees that a model would be able to differentiate between the most frequent and most basic meaning of a word.

\begin{figure}
    \centering
    \includegraphics[scale=0.35]{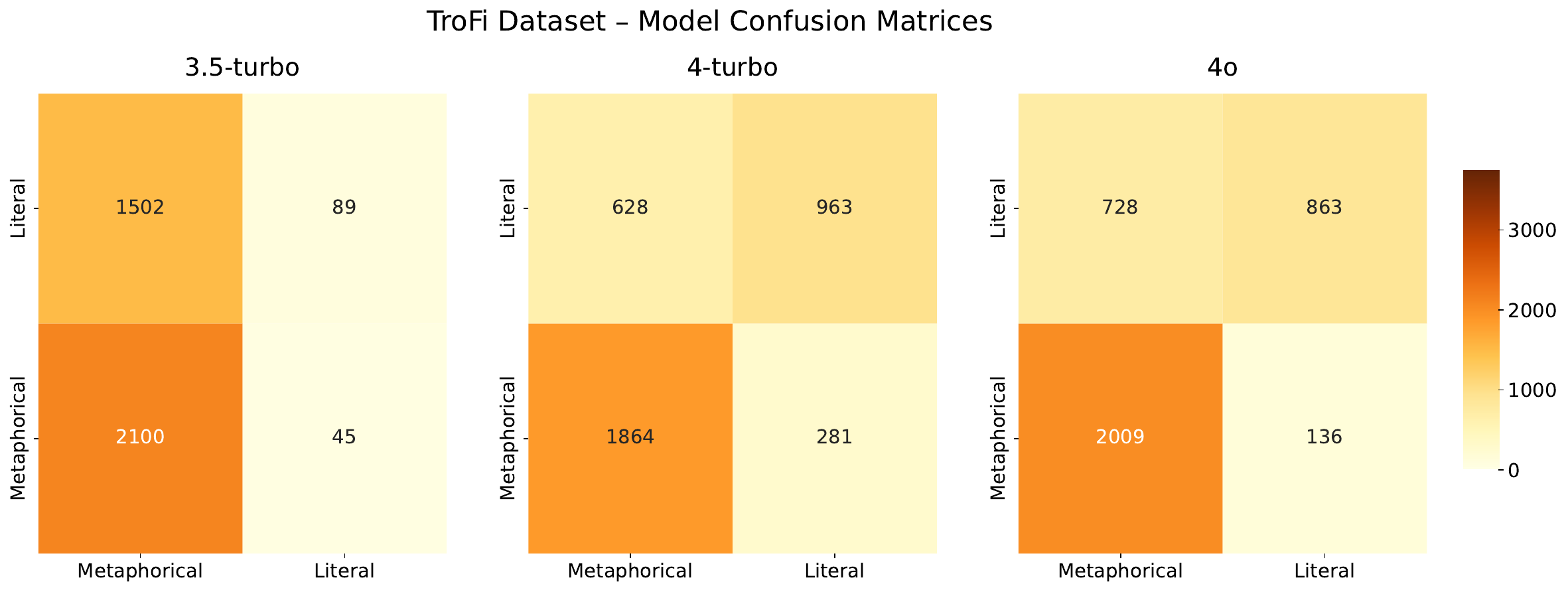}
    \caption{Confusion matrices for results on the TroFi dataset for each model. The rows represent the true values for each sample, and the columns represent the model labels.}
    \label{fig:trofi-results}
\end{figure}

We also find, however, that the models are prone to over-labeling words as metaphorical (Figure \ref{fig:trofi-results}). Whereas all three models achieve high recall for metaphor identification, they frequently label words used literally as having a more basic meaning, leading to low recall for the `literal' category and low precision for the `metaphorical' category. \texttt{3.5-turbo} particularly struggles with accurately identifying literal word usage. It is worth noting that, although there are clear cases where the models have failed to properly apply MIP, there may also be a disconnect between annotations in the TroFi corpus and what is considered metaphorical by the MIP procedure, as MIP was not used for annotation in TroFi.

\begin{figure}
    \centering
    \includegraphics[scale=0.35]{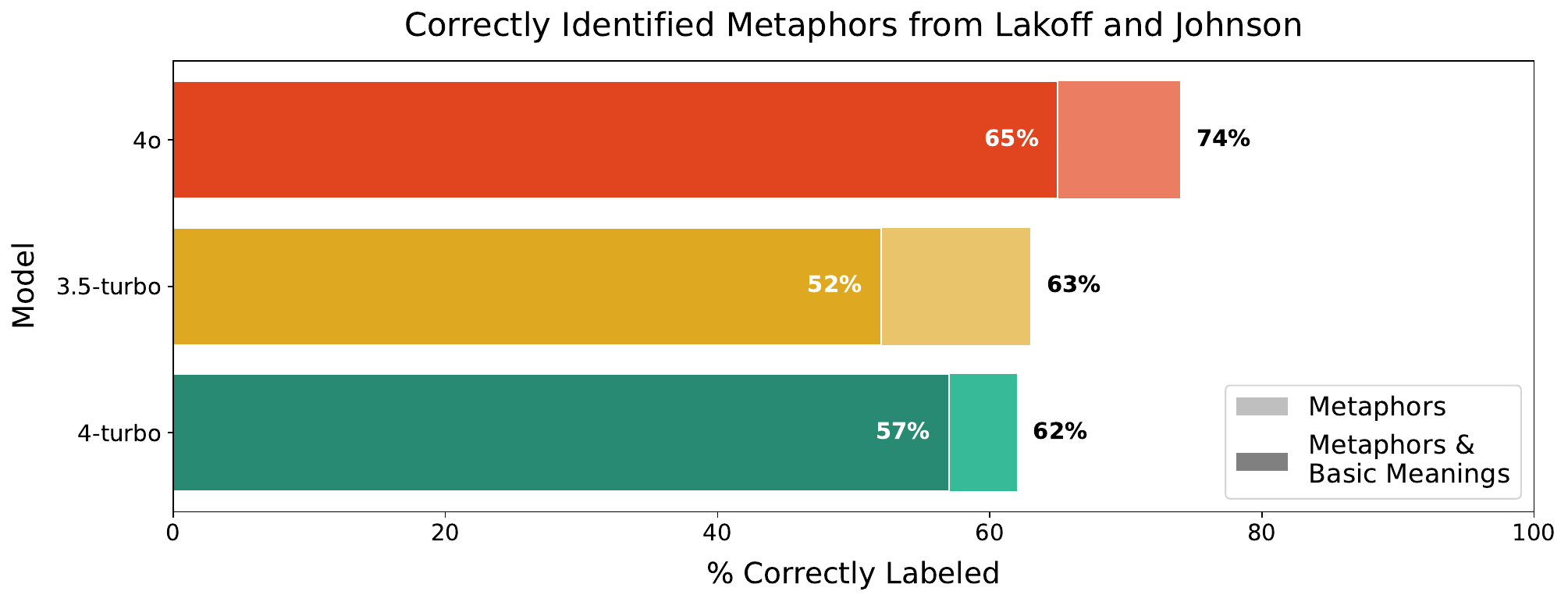}
    \caption{The percentage of samples from Lakoff and Johnson where all metaphors and basic meanings are correctly identified.}
    \label{fig:metaphor-results}
\end{figure}

\begin{figure}
    \centering
    \includegraphics[scale=0.4]{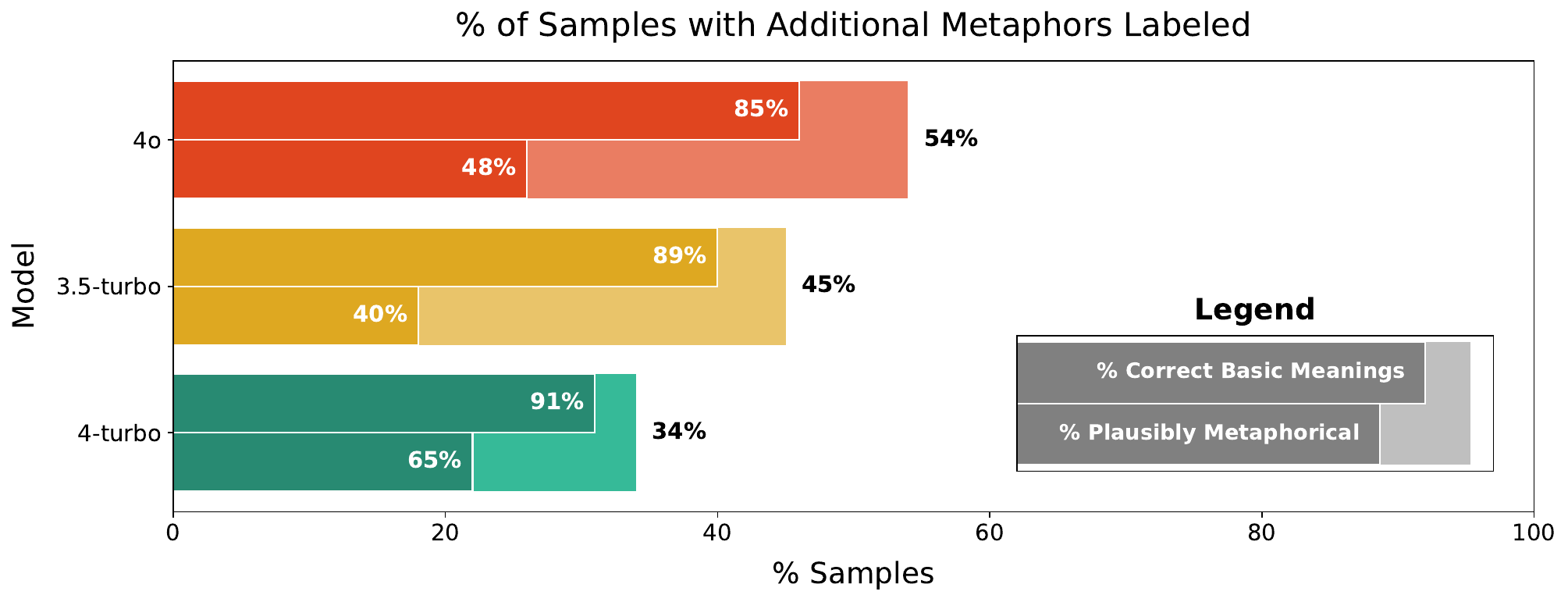}
    \caption{The percentage of samples from Lakoff and Johnson in which additional metaphors have been labeled. Smaller bars are included which represent the percentage of additional words for which the correct basic meaning is provided (upper bar) and which are plausible metaphors (lower bar).}
    \label{fig:add-metaphor-results}
\end{figure}

For the annotated subset of the MWLB dataset, all the models found all the Lakoff and Johnson metaphors in over 60\% of sentences and further provided a correct basic meaning for the metaphorical words in over 50\% of sentences (Figure \ref{fig:metaphor-results}). \texttt{4o} performed the best on this task, achieving a remarkable performance given the subtlety and complexity of many of the metaphors and the cultural knowledge required to identify both their metaphoricity and their basic meanings. \texttt{4o}'s high performance may be due in part to the fact that it was most likely to label words as metaphorical overall, having annotated additional words in the largest percentage of sentences (Figure \ref{fig:add-metaphor-results}). In only 48\% of sentences did these additional annotations identify only plausibly metaphorical words. Nonetheless, the inclusion of correct basic meanings for such a large proportion of the Lakoff and Johnson metaphors suggests that MIP was overall being correctly applied and true `knowledge' was being demonstrated, not just random chance.

\texttt{3.5-turbo} struggled the most at replicating the desired output structure, occasionally excluding words or truncating sentences, inconsistently formatting lists, and sometimes dropping the parenthetical explanations. This affected its accuracy, as the model sometimes failed to provide annotations for words central to the metaphors of interest. Additionally, \texttt{4-turbo} was least likely to annotate a word as having a more basic meaning (Figure \ref{fig:add-metaphor-results}). This meant it identified fewer Lakoff and Johnson metaphors, but a greater proportion of the additional metaphors it annotated were plausible. The basic meanings provided by \texttt{4-turbo} were usually of high quality, and tended to be more accurate and specific than those produced by \texttt{3.5-turbo}. Despite the benefits, however, this model's `caution' led it to perform worse than \texttt{4o} overall.

Each model struggled to correctly identify when smaller function words were used metaphorically, particularly prepositions like `in,' . This made the models worse at identifying so-called container metaphors, such as \textsc{life is a container} and \textsc{activities are containers}. For example, in the sentence ``That's \textit{in} the \textit{center} of my \textit{field} of vision.'', labeled by Lakoff and Johnson as \textsc{visual fields are containers}, the word `in' was overlooked as metaphorical by all three models.

The models were also challenged by metaphors in which an entity is being treated as a different kind of object, like instances of personification, \textsc{place for institution}, or \textsc{producer for product} metaphors. However, the GPT-4 models had a much greater ability to detect and provide accurate basic meanings for these. For example, in ``Let's not let Thailand become another \textit{Vietnam}.'', both \texttt{4-turbo} and \texttt{4o} correctly identified `Vietnam' as a metaphor. \texttt{4o} explained that the more basic meaning of the word ``refers to the country, whereas in this context it refers to a situation similar to the Vietnam War'' and \texttt{4-turbo} provides a similar response. Likewise, in``I hate to read \textit{Heidegger}.'', \texttt{4o} recognized that `Heidegger' is being used metaphorically and stated that the ``more basic meaning refers to the person Martin Heidegger, a German philosopher, rather than his works.'' Identifying and explaining both of these metaphors requires a nuanced understanding of both the semantics of the sentence and the cultural context surrounding them. The models cannot always perform this analysis (all three miss that `the Alamo' is metaphorical in ``Remember \textit{the Alamo}!''), but it is impressive that they are ever able to do so.

In addition, forcing the models to annotate word-by-word makes it challenging for them to identify metaphors comprising multi-word units. For example, it is difficult to determine which words should be marked as having more basic meanings in the sentence ``\textit{Get the most out of} life.'' 
Analyzing texts word-by-word can also make identifying and providing the basic meanings of metaphorical compound words more complex. For example, both `underage' and `brainchild' cause problems for \texttt{3.5-turbo}; it does not recognize that the \textit{under} of underage is metaphorical and it says that the more basic meaning of brainchild is ``a child conceived in the mind.'' In contrast, \texttt{4-turbo} says that the ``more basic meanings of `brain' and `child' are more concrete and related to physical objects or beings'' and notes that the ``more basic meaning of `under' is physically beneath something'' for underage. \texttt{4o} recognizes both words as metaphorical, but provides worse basic meanings. 

While the models are clearly fallible, they nevertheless demonstrate an impressive ability to synthesize semantic, syntactic, and cultural information. They are frequently able to recognize when a word is being used metaphorically and often provide a correct basic meaning for the word. This ability sometimes holds even for nuanced and complex metaphors.

\section{Conclusion}

We find that large, generative LMs are capable of applying the classic metaphor annotation procedure, MIP. In doing so, they demonstrate an ability to discern the ``basic meaning'' of words and thus a depth of linguistic understanding that is not obviously gleaned from language modeling pre-training. Notably, they also demonstrate that LLMs are able to execute linguistic procedures designed for human annotators. This capacity means that generative LLMs are a promising tool for large-scale computational research on conceptual metaphors, which has previously been largely infeasible. In addition, it suggests that further research on how metaphoricity is learned by models may provide insight into their ability to acquire complex linguistic knowledge that often relies on people's embodied experiences. These findings also suggest several avenues for future research, including studies into where information about words' basic meanings or metaphoricity is stored by models, an exploration of open source models ability to annotate for conceptual metaphor, and research on operationalizing other linguistic procedures for execution by LLMs.

\begin{acknowledgments}
  We would like to thank Lavinia Cerioni for her help in brainstorming this project and our colleagues at the Center for Humanities Computing for their support and advice. Additionally, part of the computation done for this project was performed on the UCloud interactive HPC system, which is managed by the eScience Center at the University of Southern Denmark.
\end{acknowledgments}

\bibliography{bibliography}

\end{document}